\documentclass[11pt,a4paper]{article}
\usepackage[hyperref]{emnlp2018}
\usepackage{times}
\usepackage{latexsym}

\usepackage{url}
\usepackage{graphicx}
\usepackage{multirow}
\usepackage{tabularx}
\usepackage{amsmath,amssymb}
\usepackage{tablists}
	\restorelistitem
\usepackage{bm}
\usepackage[group-separator={,},group-minimum-digits={3}]{siunitx}
\usepackage{CJKutf8}
\usepackage{subfigure}
\usepackage{tikz}
\usepackage{environ}
\aclfinalcopy 

\title{Sememe Prediction: Learning Semantic Knowledge from Unstructured Textual Wiki Descriptions}

\author{Wei Li, Xuancheng Ren, Damai Dai, Yunfang Wu,  Houfeng Wang,  Xu Sun\\
  MOE Key Laboratory of Computational Linguistics, \\School of Electronics Engineering and Computer Science,\\ Peking University\\ 
  {\tt liweitj47,renxc,daidamai,wuyf,wanghf,xusun@pku.edu.cn} }

\date{}

\begin{document}
\maketitle
\begin{abstract}
  Huge numbers of new words emerge every day, leading to a great need for representing them with semantic meaning that is understandable to NLP systems. Sememes are defined as the minimum semantic units of human languages, the combination of which can represent the meaning of a word. Manual construction of sememe based knowledge bases is time-consuming and labor-intensive. Fortunately, communities are devoted to composing the descriptions of words in the wiki websites. In this paper, we explore to automatically predict lexical sememes based on the descriptions of the words in the wiki websites. We view this problem as a weakly ordered multi-label task and propose a  Label Distributed seq2seq model (LD-seq2seq) with a novel soft loss function to solve the problem. In the experiments, we take a real-world sememe knowledge base HowNet and the corresponding descriptions of the words in Baidu Wiki\footnote{\url{https://baike.baidu.com/}} for training and evaluation. The results show that our LD-seq2seq model not only beats all the baselines significantly on the test set, but also outperforms amateur human annotators in a random subset of the test set. 
\end{abstract}

\section{Introduction}

With the development of the Internet, new words are emerging at an unprecedented speed. It is difficult for Natural language processing (NLP) systems to understand these new words or phrases without auxiliary information with limited contexts. Fortunately, many volunteers in the community are devoted to constructing the wiki pages for many of the new words and phrases, which makes wiki websites like Wikipedia\footnote{\url{https://en.wikipedia.org/}} and Baidu Wiki\footnotemark[1] very valuable resources. However, the descriptions in the wiki pages are depicted in natural language which are unstructured, noisy and hard for the NLP systems to understand. Therefore, there is a great need to represent these words with semantic meanings in a structured fashion that can be easily understood by the NLP systems.

\begin{table}[ht]
\centering
\small
\begin{tabular}{|p{1.9cm}|p{5cm}|}
\hline
\textbf{Word} & \begin{CJK*}{UTF8}{gbsn}缕析\end{CJK*} (analysis in detail)\\ \hline
\textbf{Description in Baidu Wiki} & \begin{CJK*}{UTF8}{gbsn}逐条认真的分析缕析行情\end{CJK*} (A careful analysis. Analyze the market in detail) \\ \hline
\textbf{Sememes} & \begin{CJK*}{UTF8}{gbsn}分析\ (analyze)\  详\ (detailed)\end{CJK*} \\ \hline

\end{tabular}
\caption{Example of Sememe Prediction via Wiki Description}\label{tab: sememe example}
\end{table}

Words can be represented with semantic sub-units from a finite set of limited size. For example, the word ``lovers'' can be approximately represented as ``\{Human $|$ Friend $|$ Love $|$ Desired\}'', the word ``\begin{CJK*}{UTF8}{gbsn}缕析\end{CJK*}'' (analysis in detail)  can be represented as ``analyze'' and ``detailed'' (see Table \ref{tab: sememe example}). Linguists define \textbf{sememes} as this kind of semantic sub-units of human languages \cite{bloomfield1926set} that express semantic meanings of concepts. This idea is similar to the idea of language universals \cite{goddard1994semantic}. 
To represent the semantic meaning of words with the sememes, researchers build sememe based knowledge bases (KBs) by annotating words with a pre-defined set of sememes. One of the usable and most well-known sememe KBs is HowNet \cite{zhendong2006hownet}. In the ontology of HowNet, there are over 2,000 sememes. They manually annotated more than 100,000 words and phrases in Chinese in a hierarchical structure. Because of its explicit way to represent knowledge (the number of sememes is limited, which embody knowledge), HowNet is easy to be adopted in NLP systems while remains understandable to human beings.

The manual construction of such KBs is very time-consuming and labor-intensive, for instance, HowNet was built for more than 10 years by a number of linguistic experts. However, many of the annotated words in the KBs are already out of date, in the meanwhile, the progress of manual construction can not catch up with the emerging speed of the new words. 

In the real world, there are many different wiki websites, such as Wikipedia,\footnote{\url{http://www.wikipedia.org}} Baidu Wiki,\footnote{\url{http://baike.baidu.com}} Hudong Wiki,\footnote{\url{http://www.baike.com}} and so on.
These websites contain millions of high-quality articles describing the world knowledge embodied in the words and phrases. For instance, Baidu Wiki contains \num{15243192} articles, mostly in Chinese. 
When people are not familiar with some words, nowadays they prefer to look up the descriptions in these wiki websites. However, for the commonly used ``classical'' words, dictionaries are still a valuable source, in which people look up the meanings of the words. 
Therefore, we think it is reasonable to use resources from both kinds of web pages.

In this paper, we intend to explore a way to predict lexical sememes of a word based on its corresponding descriptions in the wiki (dictionary) pages . We view this task as a weakly ordered multi-labeling problem (the order is already given by HowNet).

\newcite{vinyals2015order} claimed that the order between labels matters, and they proposed to use seq2seq learning for the multi-label problem. \newcite{nam2017maximizing} proposed several ways to organize the order of labels so that seq2seq would work better on the MLC task. We observe that the classical sequence-to-sequence (seq2seq) model makes a strong assumption on the order of the labels, which is not suitable for the multi-label problem. Assuming the order between tokens with heuristic rules is also problematic. Therefore, we propose a novel \textbf{label distributed seq2seq model} (LD-seq2seq) with a soft loss function to solve the problem. Since single wiki description may involve noise, and is not comprehensive, we design a multi-resource encoder that can take various description resources (e.g., descriptions from different wiki websites) into consideration.  

Our contributions lie in the following aspects:
\begin{itemize}
\item We propose to predict the sememes of a word based on its textual descriptions in wiki pages, which transforms the unstructured textual knowledge from wiki pages into distributed semantic knowledge.
\item We view this task as a weakly ordered multi-labeling problem and propose a Label Distributed Seq2seq model with a soft loss function to solve the problem. 
\item We do extensive experiments on sememe prediction and observe that our model beats all the baselines. Our model even outperforms amateur human annotators on a random subset of the test set. Furthermore, we give a detailed analysis of the error reasons with concrete examples and possible solutions. 
\end{itemize}

\section{Related Work}
HowNet has been widely used in various NLP tasks such as word similarity computation \cite{liu2002word}, word sense disambiguation \cite{duan2007word} (similar to word Clustering \cite{jin2007word}), sentiment analysis \cite{huang2014new} and name entity recognition \cite{10.1007/978-3-319-50496-4_38}. \newcite{niu2017improved} claimed that using word sememe information in HowNet can improve word representation. \newcite{zeng2018chinese} proposed to expand the Linguistic Inquiry and Word Count  \cite{pennebaker2001linguistic} lexicons based on word sememes.

\newcite{xie2017lexical} proposed to predict sememes of a word by measuring the similarity between the jointly learned word embeddings and sememe embeddings. Their solution is simple and straightforward. However, in many of the cases in real-world applications, we do not have access to the accurately learned word embeddings, especially for the new words. First, it is hard to collect enough context data for learning the embedding of new words. Second, in most of the deep learning applications, the word embeddings are fixed after training, which makes it difficult to learn the embedding of the new words and fix them into the system.

There are three main types of traditional machine learning algorithms for the Multi-Label Classification (MLC) task, problem transformation methods \cite{BR,LP,read2011classifier},algorithm adaptation methods \cite{clare2001knowledge,zhang2007ml,furnkranz2008multilabel} and ensemble methods \cite{tsoumakas2011random,szymanski2016data}. Simple neural networks models have also been applied to deal with MLC tasks \cite{zhang2006multilabel,nam2014large,benites2015haram,kurata2016improved}. 
\newcite{li2015multi} proposed to consider the previously generated labels as features for predicting new ones. \newcite{yang2018sgm} further developed this idea to use recurrent neural networks to model the correlation between labels.

\section{Our Approach}
\begin{figure*}
\begin{minipage}[t]{0.45\linewidth}
\centering
\includegraphics[width=2.8in]{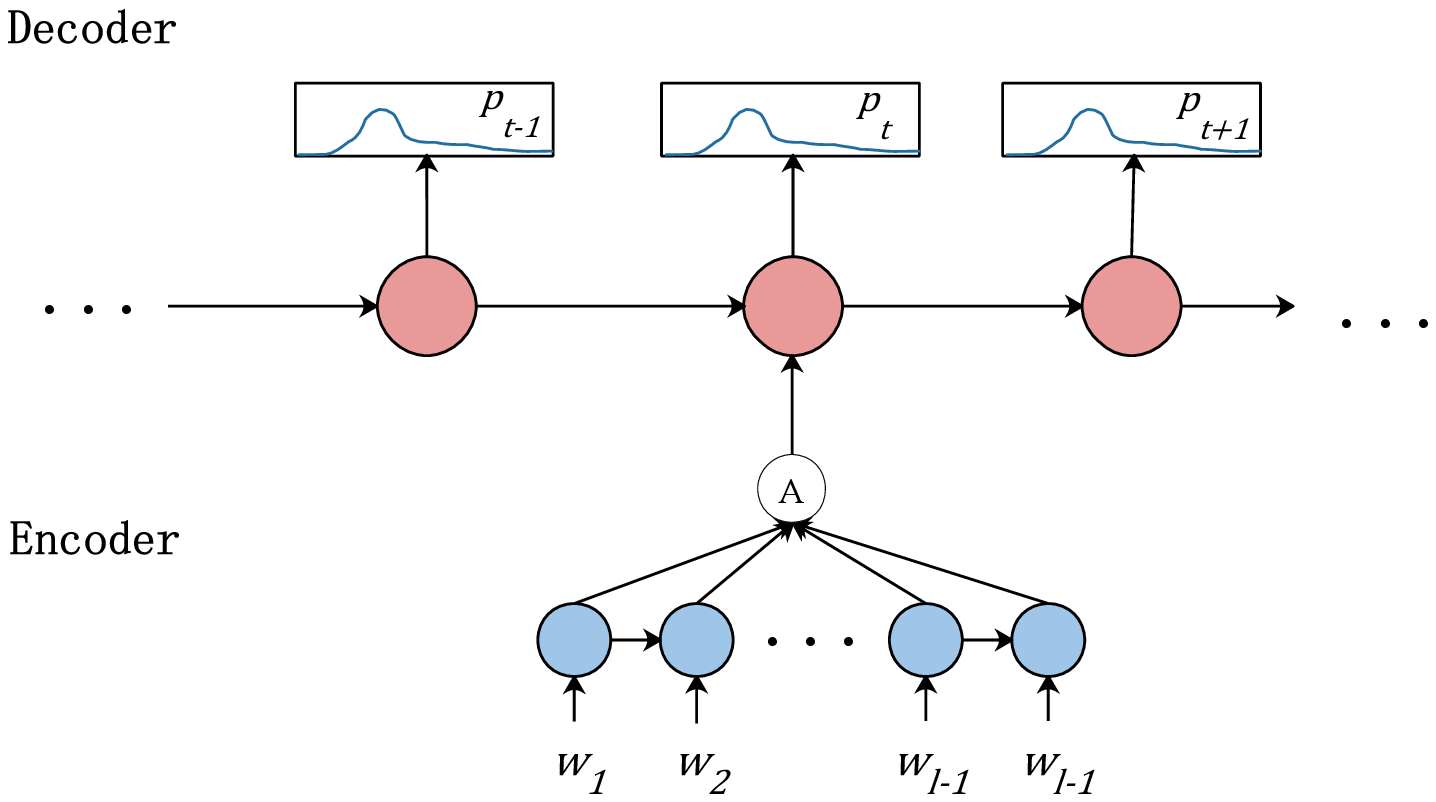}
\caption{An overview of proposed label distributed seq2seq model. We compute the loss based on a soft probability distribution rather than the one-hot distribution. 
}
\label{fig:side:a}
\end{minipage}
\hspace{0.1in}
\begin{minipage}[t]{0.52\linewidth}
\centering
\includegraphics[width=3.3in]{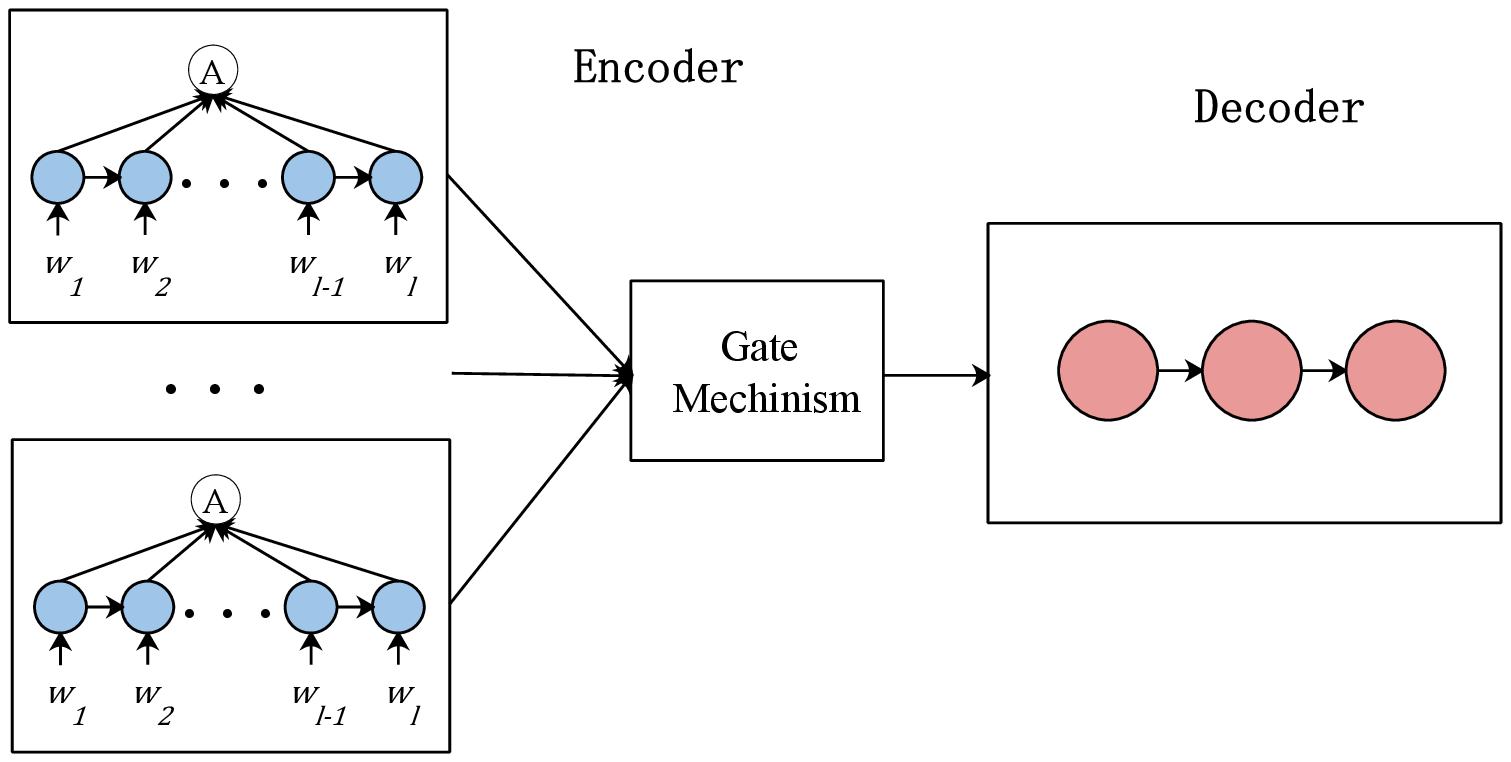}
\caption{An illustration of the multi-resource model, the different descriptions' vectors and  context vectors are combined with gate mechanism. In the figure we show two descriptions, while our model can be extended to multiple descriptions.}
\label{fig:side:b}
\end{minipage}
\end{figure*}
\label{sec:approach}
In this section, we show our solution to the sememe prediction task. An overview of our model is shown in Figure \ref{fig:side:a}.

\subsection{Task Definition}
Given one (single resource) or a few (multiple resources) textual descriptions $D = (d^{(1)},d^{(2)}, \cdots, d^{(m)})$ of a word from the wiki pages, our goal is to predict the corresponding sememes $\bm{s} = (s_1, s_2, \cdots, s_{n})$ of the word, where $\bm{s}$ is a subset of the sememe label space $S$. 
 
Our task can be modeled as finding an optimal label sequence $\bm{s^{\ast}}$ that maximizes the conditional probability $p(\bm{s} | \bm{D})$, which is calculated as follows,
\begin{equation}\label{equ:overview}
p(\bm{s} | \bm{D}) = \prod_{i=1}^{n} p(s_i|s_1, s_2, \cdots, s_{i-1}, \bm{D})
\end{equation}

\subsection{Basic Seq2seq Model for Multi-Label} \label{sec:basic seq2seq}
\newcite{vinyals2015order} proposed to use seq2seq paradigm to deal with the problem of predicting labels that form a set. They claimed that the order of the labels matters even for labels that form a set. 

\textbf{Encoder}: For one textual description $d_i$ with $l$ words in $D$, it is first encoded to $l$ hidden states $(h_1, h_2, \cdots, h_l)$ by the bidirectional gated recurrent neural networks (BiGRNN), the last of which is treated as the vector $v_d$ for the textual description,
\begin{equation}
h_t = GRU(h_{t-1}, x_t)
\end{equation}

\textbf{Decoder}: The decoder generates the sememes one by one based on the vector $v_d$. At the $t$-th time of decoding, the probability of the sememe $p_t$ is calculated as follows,

\begin{small}
\begin{eqnarray}
s_{t} = GRU([\bm{s}_{t-1} ; \bm{c}_{t} ; \bm{e}_{t-1}]) \label{equ:GRU} \\
p_{t} = softmax(\bm{W} \bm{s}_{t} + b) \\ \label{equ:predict} 
 c_t  = \sum\nolimits_{i=1}^{l}\alpha_{t,i}h_i \label{equ:context} \\
 score_{t,i}  = \bm{v}_a^T\tanh(\bm{W}_a\bm{s}_t + \bm{U}_a\bm{h}_i) \label{equ:score} \\
\alpha_{t,i}  = \frac{\exp(score_{t,i})}{\sum_{j=1}^{l}\exp(score_{t,j})}  \label{equ:alpha}
\end{eqnarray}
\end{small}
where $s_t$ is the hidden state at the $t$-th time, $c_{t}$ is the context vector calculated with the attention mechanism over the hidden states of the descriptions $(h_{1}, h_{2}, \cdots, h_{l})$, $\bm{e}_{t-1}$ is the embedding of the sememe with the highest probability predicted at the $(t-1)$-th time.

\subsection{Proposed Label Distributed Seq2seq Model}\label{sec:single-model}

We think that even though the order of the labels matters, we should not strictly restrict the order of the labels. However, the traditional cross entropy loss function applied to the classical seq2seq model actually puts a strict assumption on the order of the labels. For example, if the third token in the target sequence is predicted at the first place, it will be punished with no difference to predicting an utterly wrong token. To deal with the task of predicting weakly ordered labels (or even unordered labels), we propose a soft loss function instead of the original hard cross entropy loss function,
\begin{equation}
loss = -\sum_{i} y'_{i} log(p_i)
\end{equation}
Instead of using the original hard one-hot target probability $y_i$, we use a soft target probability distribution, which is calculated according to $y_i$ and the sememe sequence $\bm{s}$ of this sample. Let $\bm{q_s}$ denote the bag of words representation of $\bm{s}$, where only the slots of the sememes in $\bm{s}$ are filled with $1$s. We use a function $\xi$ to project the original target label probability $y$ into a new probability distribution $y'$,
\begin{equation}
y'_t = \xi(y_t, \bm{q_s})
\end{equation}
This function is designed so as to decrease the harsh punishment when the model predicts the labels in the wrong order. In this paper, we apply a simple yet effective projection function as Equation (\ref{equ:soft_loss}). It should be noted that this is an example implementation, one can also design more sophisticated projection functions if needed,
\begin{equation}
\xi(y_t,\bm{s}) = ((\bm{q_s}/M) + y_t) / 2 \label{equ:soft_loss}
\end{equation}
where $M$ is the length of $\bm{s}$. This function means that at the $t$-th time of decoding, for each target token $s_i$, we first split a probability density of $1.0$ equally across all the $M$ tokens into $1/M$. Then, we take the average of this probability distribution and the original probability $y_t$ to be the final probability distribution at time $t$.

\subsection{Multi-Resource Model}\label{sec:encoder}
Description resource from a single source can be unreliable and is not able to express the meaning of the word comprehensively.
In this paper, we propose to use a multi-resource encoder to make use of descriptions from multiple resources. An overview of this model is shown in Figure \ref{fig:side:b}. To demonstrate the effectiveness of multiple resources, we implement our encoder using two resources for simplicity,  but it can be extended to more resources without much effort. 

Assume for a word, we have two textual descriptions $d^{(1)}$ and $d^{(2)}$, containing $l^{(1)}$ and $l^{(2)}$ words $(w_{1}^{(1)}, w_{2}^{(1)}, \cdots, w_{l^{(1)}}^{(1)})$ and  $(w_{1}^{(2)}, w_{2}^{(2)}, \cdots, w_{l^{(2)}}^{(2)})$ respectively. We use BiGRNN to encode the two descriptions separately into two sequences of hidden states $(h_{1}^{(1)}, h_{2}^{(1)}, \cdots, h_{l^{(1)}}^{(1)})$ and $(h_{1}^{(2)}, h_{2}^{(2)}, \cdots, h_{l^{(2)}}^{(2)})$. We use the hidden states at the last time step $h_{l^{(1)}}^{(1)}$ and $h_{l^{(2)}}^{(2)}$  as the representation for the corresponding descriptions $d^{(1)}$ and $d^{(2)}$, which we denote as $v_{d}^{(1)}$ and $v_{d}^{(2)}$.

To combine the two vectors $v_{d}^{(1)}$ and $v_{d}^{(2)}$ into one uniform $v_{d}$, we apply the gate mechanism, which is calculated as follows,
\begin{small}
\begin{eqnarray}
g_{1} = \sigma(W_{1} [v_d^{(1)} ; v_d^{(2)}] + b_1) \\
g_{2} = \sigma(W_{2} [v_d^{(1)} ; v_d^{(2)}] + b_2) \\
v_{d} = g_{1} \odot v_{d}^{(1)} + g_{2} \odot v_{d}^{(2)}
\end{eqnarray}
\end{small}
where $\sigma$ indicates the $sigmoid$ function, $W_{1}$, $W_{2}$, $b_1$ and $b_2$ are learnable parameters, $\odot$ means the element-vise multiplication.

The decoder part follows the same structure as the model in Section \ref{sec:single-model}, except that we first separately calculate the context vectors $c_t^{(1)}$ and $c_t^{(2)}$ with attention mechanism. Then we use gate mechanism to combine the two vectors $c_t^{(1)}$ and $c_t^{(2)}$ into one context vector $c_{t}$. The gate mechanism here follows the same process for the combination of $v_d^{(1)}$ and $v_d^{(2)}$ with different parameters.

\section{Experiment}
\label{sec:experiment}
\subsection{Dataset}
\textbf{HowNet}: HowNet is a knowledge base that uses sememes to represent the semantic meaning of a word or a phrase. There are over \num{100000} annotated words in HowNet. Words can have multiple senses. Each sense is further represented by a combination of no more than 8 sememes. The sememes form a hierarchical structure. However, following the settings of most of the previous work, we do not consider the specific relations between sememes, but only consider the order between them, which we call weakly ordered sememes. For simplicity, we do not consider multiple senses, and just assume that the first sense of the word is its basic sense. 

\noindent \textbf{Wiki Pages}: 
Because the words annotated in the HowNet consist of both common words and newly emerged words (by that time), we choose two description sources for the words, Baidu Wiki\footnote{\url{http://baike.baidu.com}} (\begin{CJK*}{UTF8}{gbsn}百度百科\end{CJK*}) and Baidu Dictionary\footnote{\url{http://dict.baidu.com}} (\begin{CJK*}{UTF8}{gbsn}百度词典\end{CJK*}). Baidu Wiki contains 15,244,702 articles that are edited by the volunteers with a lot of new emerged words, while Baidu Dictionary is similar to language dictionaries (still from crowd-source) with better quality definitions and descriptions for common words.

We get the textual descriptions of the words annotated in the HowNet from Baidu Wiki and Baidu Dictionary, and get \num{62810} words that have attached descriptions (if at least one of the descriptions from two sources exist, it is counted as one case). We randomly split the data into three parts, Train (80\%), Dev (10\%) and Test (10\%).

\subsection{Baseline Models}
\begin{itemize}
\setlength{\itemsep}{1.5pt}
\setlength{\parskip}{1.5pt}
\item ML-KNN (Multi-label KNN): This is the k-Nearest Neighborhood classification method adapted to multi-label classification. 
\item LP (Label Powerset): LP \cite{LP} is a problem transformation approach to multi-label classification that transforms a multi-label problem to a multi-class problem with one multi-class classifier trained on all unique label combinations found in the training data. 
\item CC (Classifier Chain): For the label space with $L$ labels, CC \cite{read2011classifier} trains $L$ classifiers ordered in a chain according to the Bayesian chain rule. 
\item BR (Binary Relevance): BR \cite{BR} transforms a multi-label classification problem with $L$ labels in the label space into $L$ single-label separate binary classification problems using the same base classifier. 
\item RNN-MLLR (RNN multi-label logistic regression): This model uses the same multi-resource encoder of our proposed model, while uses the one-versus-all logistic regression multi-label classifier to predict the sememes based on the encoded vector of the descriptions.
\end{itemize}
\subsection{Experiment Details}
For the textual descriptions, we use characters as the input, the vocabulary size of characters is \num{11097}. We randomly initialize the character embeddings. There are \num{2185} sememes in the HowNet. We use word2vec \cite{mikolov2013distributed} toolkit to pre-train the embeddings of the sememes with default parameters of the code to capture the co-occurrence relationship of the sememes. The embeddings of the sememes are fine-tuned during training. The dimension of both the character embeddings and sememe embeddings are 200. 
All the dimensions of hidden states are set to 300. The batch size is 20. $<EOS>$ token is added to the end of a sememe sequence to indicate when to stop prediction. We use Adam optimizer \cite{KingmaBa2014} to minimize the loss. We train our model for 10 epochs, and choose the model parameters from the epoch that gets the highest F1 score on the Dev set.

\subsection{Results and Analysis}
We use micro Precision (P), Recall rate (R) and F1 score as the evaluation metrics. 

\begin{table}[tb]
\centering
\begin{tabular}{|l|ccc|}
\hline
Model & P & R & F1  \\ \hline
ML-KNN & 29.34 & 9.26 & 14.08 \\
LP    & 26.06 & 23.92 & 24.94 \\
BR    & 32.30 & 21.59 & 25.88\\
CC    & 33.33 & 21.37 & 26.04\\
RNN-MLLR & 44.26 & 33.54 & 38.16 \\ \hline
Basic Seq2seq & 43.86 & 40.92 & 42.34 \\
LD-Seq2seq (Proposal) &  47.96 & 41.99 & 44.78 \\ \hline
\end{tabular}
\caption{Comparison with different baseline models. All the models use two wiki resources in this table. ``P'' means Precision,``R'' means recall rate.}\label{tab:overall results}
\end{table}

\noindent \textbf{Comparison with Baselines:} In Table \ref{tab:overall results}, we show our experiment results compared with the baseline methods. From the results we can see that clustering based method ML-KNN performs the worst for sememe prediction. We assume that this is because the textual descriptions are very diverse, which makes KNN hard to determine the borders among space of different labels. Methods that aim to transform classifiers to multi-label task perform closely to each other, with F1 scores around 25\%. 

Compared with traditional machine learning methods (ML-KNN, LP, CC, BR), neural network based methods (RNN-MLLR, Basic seq2seq) performs much better, which beats other baselines by a big margin. Although RNN-MLLR achieves good results, it is still not as good as seq2seq based model. We assume that this is because MLLR based models are not very good at modeling the connections between labels. In our sememe prediction task, the sememes are in weak order. Moreover, some sememes are strongly related to some others and some sememes often co-occur. For instance, when the sememe ``Emotion'' occurs, it is likely to be followed by ``FeelingByBad'', ``generic'' and ``desired''. Our proposed Label Distributed seq2seq model gets the best performance, we assume that this is because even though order between labels matters \citep{vinyals2015order}, for the weakly ordered multi-label problem, a strong assumption on ordering hurts the performance, and our soft loss function can effectively relieve the problem.

\begin{table}[tb]
\centering
\begin{tabular}{|l|ccc|}
\hline
Method & Precision & Recall & F1 \\ \hline
Human &  21.89 & 57.36 & 31.69 \\
Human+Wiki & 23.62 & 62.79 & 34.32 \\
\hline
Proposal & 53.92 & 42.64 & \textbf{47.62} \\ \hline
\end{tabular}
\caption{Comparison with Human performance on a random subset of test samples. \textit{Human} means that the annotator does not have access to the wiki descriptions. \textit{Human+Wiki} means the annotator has access to the wiki descriptions.}\label{tab:human performance}
\end{table}

\vspace{.1in}
\noindent \textbf{Comparison with Human Performance:} In Table \ref{tab:human performance}, we show the results of amateur human and our model's result on a subset of the test set. We randomly select \num{100} samples from the test set, and ask human annotators to select $1 \sim 5$ sememes out of 20 that they think can describe the meaning of the word. Because the annotators do not have background knowledge on HowNet, the annotation task is actually simpler than annotating from scratch. The annotators are highly educated (with proper knowledge) amateur native speakers without special training on linguistics or the annotation system of HowNet. We guarantee that all the correct sememes are within the selected 20 sememes. The annotators are asked to first predict the sememes based on their common sense (Human), then they are provided with the descriptions from Baidu Wiki and asked to do the work again (Human - Wiki). 

From the results we can see that even for human beings, it is hard to predict the sememes completely right without special training on the annotation system of HowNet. Human annotators are able to understand the semantic meaning of the word and can understand the description very well. However, they tend to predict more sememes than there actually are, which is reflected by the high recall rate. The imbalance between precision and recall indicates that the sememe architecture of HowNet may have the problem of being too fine-grained, many sememes other than the actual ones are also related to the word, meaning wise. Still, by referring to wiki descriptions, human annotators are able to predict more precisely, this is because there are some rare words or entities in the dataset that people seldom use in the real life. Although the recall rate of our proposed model is not as high as human annotators, its precision beats human annotators by a big margin, which makes the F1 score higher than human. We assume that this is because by learning from the big bulk of training data, our model is more likely to be consistent with the logic of the annotation system.
 
\vspace{.1in}
\noindent \textbf{Effect of Proposed Soft Loss Function:}  From Table \ref{tab:overall results} we can see that seq2seq model with our novel soft loss (LD-Seq2seq) performs much better than the basic seq2seq model. We think that this is because our novel loss function eases the restriction on the order between labels. 
For example, assume the target sememes are $(s_1,s_2,s_3)$ in order. At the first time step of decoding, the one-hot loss function would strongly punish the decoder from giving $s_2$ or $s_3$ probabilities, which may confuse the decoder, because at the moment the difference between time step 1 and time step 2 may not be significant when the order of the labels are not obvious. However, our soft loss function would still lead the decoder to firstly choose $s_1$, while the two labels $s_2$ and $s_3$ are also encouraged with some probability less than $s_1$. The experiment results show that this modification is very effective to make seq2seq work well on the multi-label problem.

%

\vspace{.1in}
\noindent \textbf{Effect of Applying Multi-Resource:} From Table \ref{tab:Encoder results} we observe that using multiple resources instead of a single one can greatly improve the performance. This corresponds with our expectation as more descriptions can provide more comprehensive information of the word from various aspects. Moreover, since the alignment between sememes and descriptions are noisy, the gate mechanism can automatically decide how much one description contributes to the prediction based on its relatedness. Between the two resources we used (Baidu Wiki and Baidu Dictionary), dictionary-style resource provides much higher precision (47.15 $\sim$ 42.91), we assume this is because the descriptions in this kind of resource have better quality in general. However, many new words and rare words are not included in the dictionary and some of the entries in the Baidu Dictionary have noisy descriptions as well (e.g., English descriptions instead of Chinese), so dictionary alone does not predict as well as the multi-resource one.

\begin{table}[t!]
\centering
\begin{tabular}{|l|ccc|ccc|}
\hline
Model & Precision & Recall & F1  \\ \hline
SingleRes-Wiki & 42.91 & 27.75 & 33.70, \\
SingleRes-Dict & 47.15 & 29.83 & 36.54 \\
MultiRes & 47.96 & 41.99 & \textbf{44.78} \\
\hline
\end{tabular}
\caption{Results of using different resources. The seq2seq model applies the basic architecture without adaptation to the multi-label problem. \textit{SingleRes} indicates that the encoder only considers a single textual resource. \textit{MultiRes} indicates that the encoder considers multiple textual resources (Wiki and Dictionary).}\label{tab:Encoder results}
\end{table}
 
\subsection{Error Analysis and Case Study}
\begin{figure*}[t!]
\begin{minipage}[t]{0.5\linewidth}
\centering
\def\angle{0}
\def\radius{3}
\def\cyclelist{{"red","green","red!40!green!60!","red!60!green!40!","pink", "orange"}}
\newcount\cyclecount \cyclecount=-1
\newcount\ind \ind=-1
\begin{tikzpicture}[nodes = {font=\sffamily \small},scale=0.5]
  \foreach \percent/\name in {
      29/Wrong,
      24/Correct,
      24/Plausible,
      23/Partial,
    } {
      \ifx\percent\empty\else               
        \global\advance\cyclecount by 1     
        \global\advance\ind by 1            
        \ifnum5<\cyclecount                 
          \global\cyclecount=0              
          \global\ind=0                     
        \fi
        \pgfmathparse{\cyclelist[\the\ind]} 
        \edef\color{\pgfmathresult}         
        \draw[fill={\color!50!white},draw={\color}] (0,0) -- (\angle:\radius)
          arc (\angle:\angle+\percent*3.6:\radius) -- cycle;
        \node at (\angle+0.5*\percent*3.6:0.7*\radius) {\percent\,\%};
        \node[pin={[pin distance=5pt]\angle+0.5*\percent*3.6:\name}]
          at (\angle+0.5*\percent*3.6:\radius) {};
        \pgfmathparse{\angle+\percent*3.6}  
        \xdef\angle{\pgfmathresult}         
      \fi
    };
\end{tikzpicture}
\caption{The distribution of prediction result types.}
\label{fig:pie chart result type}
\end{minipage}
\hspace{0.1in}
\begin{minipage}[t]{0.5\linewidth}
\centering
\def\angle{0}
\def\radius{3}
\def\cyclelist{{"red","red!50!orange!50!","yellow", "green","blue","blue!50!red!50!"}}
\newcount\cyclecount \cyclecount=-1
\newcount\ind \ind=-1
\begin{tikzpicture}[nodes = {font=\sffamily \small},scale=0.5]
  \foreach \percent/\name in {
  	  24.14/Literal,	
  	  20.69/Close,		
      17.24/Unable,		
      17.24/Polysemy,	
      10.34/Complex,	
      6.90/Pattern,		
      3.45/Too Simple	
    } {
      \ifx\percent\empty\else               
        \global\advance\cyclecount by 1     
        \global\advance\ind by 1            
        \ifnum5<\cyclecount                 
          \global\cyclecount=0              
          \global\ind=0                     
        \fi
        \pgfmathparse{\cyclelist[\the\ind]} 
        \edef\color{\pgfmathresult}         
        \draw[fill={\color!50},draw={\color}] (0,0) -- (\angle:\radius)
          arc (\angle:\angle+\percent*3.6:\radius) -- cycle;
        \node at (\angle+0.5*\percent*3.6:0.7*\radius) {\percent\,\%};
        \node[pin={[pin distance=5pt]\angle+0.5*\percent*3.6:\name}]
          at (\angle+0.5*\percent*3.6:\radius) {};
        \pgfmathparse{\angle+\percent*3.6}  
        \xdef\angle{\pgfmathresult}         
      \fi
    };
\end{tikzpicture}
\caption{The distribution of error types in ``Wrong''.}
\label{fig:pie chart error type}
\end{minipage}
\end{figure*}

\begin{table*}[t!]
\small
\centering
\begin{tabular}{|p{3cm}|p{5cm}|p{5cm}|c|}
\hline
Word & Reference & Prediction & Category\\
\hline
\begin{CJK*}{UTF8}{gbsn}历史唯物主义\  (historical materialism)\end{CJK*} & \begin{CJK*}{UTF8}{gbsn}知识\ (knowledge),\ 思想\ (thinking),\ 物质\ (physical),\ 主\ (primary),\ 最\ (most)\end{CJK*} & \begin{CJK*}{UTF8}{gbsn}知识\ (knowledge),\ 思想\ (thinking),\ 物质\ (physical),\ 主\ (primary),\ 最\ (most)\end{CJK*} & Correct \\ \hline
\begin{CJK*}{UTF8}{gbsn}宦门\  (official family)\end{CJK*} & \begin{CJK*}{UTF8}{gbsn}家庭\ (family),\ 人\ (human),\ 官\ (official)\end{CJK*} & \begin{CJK*}{UTF8}{gbsn}家庭\ (family),\ 官\ (official)\end{CJK*} & Partial\\ \hline
\begin{CJK*}{UTF8}{gbsn}混纺\end{CJK*} (blend fabric) &  \begin{CJK*}{UTF8}{gbsn}人工物\ (artifact),\ 衣物\ (clothing),\  用具\ (tool)\end{CJK*}  & \begin{CJK*}{UTF8}{gbsn}材料\ (material),\ 衣物\ (clothing),\ 用具\ (tool)\end{CJK*} & Plausible \\ \hline
\begin{CJK*}{UTF8}{gbsn}国有化\end{CJK*} (nationalize) & \begin{CJK*}{UTF8}{gbsn}变性态\ (ize),\ 归属中央\ (central)\end{CJK*} & \begin{CJK*}{UTF8}{gbsn}地方\ (place),\ 有\ (own),\ 国家\ (country),\ 政\ (politics)\end{CJK*} & Wrong \\ \hline
\end{tabular}
\caption{Examples of word and sememes. \textit{Reference} indicates the standard sememes in HowNet, \textit{Prediction} indicates our predicted results. The categories of the examples are corresponding to Figure \ref{fig:pie chart result type}.}\label{tab:case study}
\end{table*}

In Figure \ref{fig:pie chart result type}, we show the distribution of the results from a randomly chosen subset of test samples (100 samples) and give some concrete examples of the sememe prediction in Table \ref{tab:case study}. We use accuracy (the case is viewed as right only if all of its sememes are matched) as the evaluation metric in the error analysis. ``\textit{correct}'' means the prediction is completely right.

In Figure \ref{fig:pie chart result type}, ``\textit{Wrong}'' means that our model makes wrong predictions. For instance, for the word  \begin{CJK*}{UTF8}{gbsn}国有化\end{CJK*} (nationalize), the standard answer is ``\textit{-ize}'' and ``\textit{central}'', while our prediction is ``place'',``own'',``country''  and ``politics'', none of the predicted sememes are in the answer set, but these sememes actually make sense, because ``nationalize '' is indeed to make something ``own'' by the ``country'', which is usually an action of ``politics'', our prediction fails to capture the dynamic procedure of ``-ize'', but still this sequence of sememes can describe some aspects of the word, thus being able to help in downstream tasks.

``\textit{Partial}'' means that part of the result is correct or the result is a subset of the real answer, for instance, for the word ``\begin{CJK*}{UTF8}{gbsn}\textit{宦门}\end{CJK*}\textit{ (official family)}'', our prediction is ``\textit{family}'' and ``\textit{official}'', while the correct answer is ``\textit{family}'', ``\textit{human}'' and ``\textit{official}'', our prediction captures most part of the meaning, and the missing sememe ``human'' can actually be deduced by the sememe ``family''.  

``\textit{Plausible}'' means that we think the predicted sememes can also reflect the meaning of the word or better, even different from the original ones, for example, for the word ``\begin{CJK*}{UTF8}{gbsn}\textit{混纺}\end{CJK*} \textit{(blend fabric)}'', our prediction is ``\textit{material}'', ``\textit{clothing}'' and ``\textit{tool}''  while the answer is ``\textit{artifact}'', ``\textit{clothing}'' and ``\textit{tool}''. The difference between two sequence of sememes lie between ``material'' and ``artifact'', blend fabric is clearly an artificial material, both the answer and our prediction captures one aspect of the word, our sequence of sememes are even better for presenting the semantic meaning of the word. The existence of plausible predictions (not entirely equal to the reference) may be related to the annotation system of HowNet. Some of the sememes we observe in the reference are very sparse, for instance, ``weatherFine'' is a sememe in HowNet, which we think can be split into other sememes like ``weather'' and ``beGood''.

Except for the wrong predictions (\textit{29\%}), we observe that the rest of the prediction result types are all similar to or can be substitution to the standard sememes of the word. We think for these parts of the predictions, the predicted sememes are able to  represent most part of the meaning of the word, which is helpful for downstream tasks. Actually, even part of the wrong predictions can be of help, which we will explain in detail. 

In Figure \ref{fig:pie chart error type}, we further split the reason of the ``Wrong'' predictions in Figure \ref{fig:pie chart result type} into \textbf{seven} categories. 

\noindent\textbf{Literal:} Among the reasons, a large part (``\textit{Literal}'' \textit{24.14\%}) is because the model is distracted by the literal meaning of some part of the descriptions that is not the key information about the word. For example, for the word ``\begin{CJK*}{UTF8}{gbsn}\textit{磕}\end{CJK*} (\textit{knock})'', our model predicts the sememes ``\textit{position}'' and ``\textit{wholly}'', because there are expressions about position like ``\begin{CJK*}{UTF8}{gbsn}碰在硬东西\textbf{上}\end{CJK*}'' (knocked \textit{\textbf{on}} a hard thing), ``\begin{CJK*}{UTF8}{gbsn}人与人\textbf{之间}\end{CJK*}'' (\textit{\textbf{between}} people) and ``\begin{CJK*}{UTF8}{gbsn}使附着物\textbf{掉下来}\end{CJK*}'' (make the attachment \textit{\textbf{off}}), these expressions are all concerned about the \textit{\textbf{position}} of something, which mislead the model.

\noindent\textbf{Close:} \textit{20.69\%} of the wrong predictions are actually close to the answers. ``\begin{CJK*}{UTF8}{gbsn}国有化\end{CJK*}'' (nationalize) we mentioned above is an example of this type.

\noindent\textbf{Polysemy:} 17.24\% of the wrong predictions are because of polysemy, that is, some words have multiple meanings, the standard sememes refers to a different meaning from the description. For example, ``\begin{CJK*}{UTF8}{gbsn}一如\end{CJK*}'' can mean ``title of a rank in karate'' or ``the same'', the sememes refer to the meaning of ``the same'', while the description in the wiki is about karate. The mismatch between the description and the answer causes such problems. 

\noindent\textbf{Complex:} \textit{10.34\%} of the wrong predictions are because the descriptions are too complex or long, which usually include many other meanings of the word. Because we only use a heuristic way to align the senses with the description, and the senses in the descriptions of the wiki are not clearly aligned, sometimes the sememes in the reference is only a part of the description, which is not in the dominant position.  For example, the word ``\begin{CJK*}{UTF8}{gbsn}践履\end{CJK*}'' can mean ``step on'' and ``fulfill'', ``step on'' is the original meaning of the word, however, the most common usage of this word now points to the meaning of ``fulfill''. In the description, a large part is describing the meaning ``step on'' and giving instances of this meaning. This makes our model focus on the wrong part of the description, thus making wrong predictions.

\noindent\textbf{Pattern:} \textit{6.9\%} of the wrong predictions are because the pattern of the annotated answer, most of which are involved with the explanation of some rarely used Chinese characters. For example, the word ``\begin{CJK*}{UTF8}{gbsn}轲\end{CJK*}'' means ``wooden vehicle'', but this original meaning is rarely used now, and the word is more acknowledged as part of the name of a saint in China ``\begin{CJK*}{UTF8}{gbsn}孟轲\end{CJK*}'' (Mencius), so the sememes in the reference are ``character'' and ``China''. 

\noindent\textbf{Too Simple:} 3.45\% of the wrong predictions are because the descriptions from the wiki are too simple. For example, the description of the word ``\begin{CJK*}{UTF8}{gbsn}猛子\end{CJK*}'' is ``\begin{CJK*}{UTF8}{gbsn}扎猛子\end{CJK*}'', which is just another way of expression without much explanation.

\noindent\textbf{Unable:} We can not tell why our model fails to predict the right answer for the rest of the wrong predictions (17.24\%). Under this circumstance, the descriptions are clear, but the predicted sememes are not concerned about the description.

To solve the mistakes we mention above, several possible methods can be applied. First, a more powerful word sense alignment step can be applied, this can make the description and the sememes correspond to each other. Second, the annotation system can be modified, so that the sparsity of the sememes can be reduced and less overlapped. Third, context of the words can be introduced to help distinguish between different senses.

\section{Conclusion and Future Work}
\label{sec:conclusion}
In this paper, we focus on the task of learning knowledge from unstructured textual descriptions from wiki pages. We choose to represent words and phrases with weakly ordered sememes. To predict the sememes of a word based on the descriptions, we propose to apply a seq2seq based model. We observe that directly applying seq2seq framework is problematic because of its strong assumption on the order between labels. To make seq2seq model more suitable for multi-label tasks,  we propose a novel soft loss function that turns the one-hot target label into a probability distribution. To make prediction more accurate, we also propose a multi-resource encoder that makes use of multiple wiki resources. Experiment results show our label distributed seq2seq model works well on the sememe prediction task. The performance is even better than amateur human on a randomly selected subset of the test set. We make a detailed error analysis and propose possible solutions.

In the future, we would like to explore how to better align the word senses with the articles in the wiki pages. It would also be interesting to take the more sophisticated structures of sememes into consideration.

\bibliography{emnlp2018}
\bibliographystyle{acl_natbib_nourl}

\end{document}